\documentclass[journal]{IEEEtran}

\usepackage{amsmath,amsfonts,bm}

\def\eqref#1{equation~\ref{#1}}

\def\1{\bm{1}}

\DeclareMathAlphabet{\mathsfit}{\encodingdefault}{\sfdefault}{m}{sl}
\SetMathAlphabet{\mathsfit}{bold}{\encodingdefault}{\sfdefault}{bx}{n}

\usepackage{hyperref}
\usepackage{threeparttable}
\usepackage{booktabs}
\usepackage{multirow}
\usepackage{graphicx}
\usepackage{algorithm}
\usepackage{algorithmic}
\usepackage{booktabs}
\usepackage{color}
\usepackage{url}

\floatname{algorithm}{Algorithm}

\begin{document}

\title{Boosting ship detection in SAR images with complementary pretraining techniques}
%
%
%

\author{Wei Bao, Meiyu Huang, Yaqin Zhang, Yao Xu, Xuejiao Liu, Xueshuang Xiang

\thanks{Wei Bao is with Qian Xuesen Laboratory of Space Technology, China Academy of Space Technology, Beijing, China; School of Information and Electronics Engineering, Beijing Institute of Technology, Beijing,
China (e-mail: baowei97@163.com).}

\thanks{Meiyu Huang, Yao Xu, Xuejiao Liu, Xueshuang Xiang are with Qian Xuesen Laboratory of Space Technology, China Academy of Space Technology, Beijing, China (e-mail: huangmeiyu@qxslab.cn; xuyao@qxslab.cn; liuxuejiao@qxslab.cn; xiangxueshuang@qxslab.cn).}

\thanks{Yaqin Zhang is with Qian Xuesen Laboratory of Space Technology, China Academy of Space Technology, Beijing, China; School of Mathematics and Computational Science, Xiangtan University, Xiangtan, China (e-mail: 493834755@qq.com).}
\thanks{Corresponding author: Meiyu Huang, Xueshuang Xiang.}
\thanks{This work is supported by the Beijing Nova Program of Science and Technology under Grant Z191100001119129 and the National Natural Science Foundation of China 61702520.}
}

\maketitle

\begin{abstract}
Deep learning methods have made significant progress in ship detection in synthetic aperture radar (SAR) images. The pretraining technique is usually adopted to support deep neural networks-based SAR ship detectors due to the scarce labeled SAR images. However, directly leveraging ImageNet pretraining is hardly to obtain a good ship detector because of different imaging perspective and geometry. In this paper, to resolve the problem of inconsistent imaging perspective between ImageNet and earth observations, we propose an optical ship detector (OSD) pretraining technique, which transfers the characteristics of ships in earth observations to SAR images from a large-scale aerial image dataset. On the other hand, to handle the problem of different imaging geometry between optical and SAR images, we propose an optical-SAR matching (OSM) pretraining technique, which transfers plentiful texture features from optical images to SAR images by common representation learning on the optical-SAR matching task. Finally, observing that the OSD pretraining based SAR ship detector has a better recall on sea area while the OSM pretraining based SAR ship detector can reduce false alarms on land area, we combine the predictions of the two detectors through weighted boxes fusion to further improve detection results. Extensive experiments on four SAR ship detection datasets and two representative CNN-based detection benchmarks are conducted to show the effectiveness and complementarity of the two proposed detectors, and the state-of-the-art performance of the combination of the two detectors. The proposed method won the sixth place of ship detection in SAR images in 2020 Gaofen challenge.

\end{abstract}


%
\IEEEpeerreviewmaketitle

\section{Introduction}
%
%
%
%
\IEEEPARstart{S}{ynthetic} Aperture Radar (SAR) is an active microwave remote sensing imaging radar with the capability of targeting objects in all-day and all-weather conditions, and has been widely applied in many military and civil fields. Ship detection in high-resolution SAR images has drawn considerable attention for its broad application prospects, such as marine surveillance~\cite{cerutti2008wide}, military intelligence acquisition~\cite{brusch2010ship} and so on. Many traditional SAR ship detection methods have been proposed~\cite{hou2014multilayer,gao2016scheme,zhu2016projection} to detect multi-scale ships in complex surroundings. As one of the most commonly used techniques, the constant false-alarm rate (CFAR) method~\cite{hou2014multilayer}, adaptively adjusts the threshold given a false-alarm rate and leverages the estimated statistical distributions to distinguish objects from the background with the calculated threshold. However, traditional detection methods suffer from tremendous difficulties in accurate detection due to weak feature extraction capabilities.

Recently, benefited from the rapid development of deep learning, remarkable breakthroughs have been made in deep convolutional neural networks (CNN)~\cite{sainath2013deep}-based detection methods. Generally, CNN-based detection methods can be divided into two categories: two-stage detection methods, such as Faster R-CNN~\cite{ren2015faster}, Mask R-CNN~\cite{he2017mask}, Cascade R-CNN~\cite{cai2018cascade}; and one-stage detection methods, such as SSD~\cite{liu2016ssd}, RetinaNet~\cite{lin2017focal}, YOLO~\cite{Redmon_2016_CVPR}. Specifically, two-stage detection methods adopt the feature maps generated from backbone networks, e.g. Residual Network (ResNet)~\cite{he2016deep}, to preliminarily extract class-agnostic region proposals of the potential objects with negative locations filtered out, and then further refine these proposals and classify them into different categories. Different from two-stage detection methods, one-stage detection methods omit the region proposals generation process and consider object detection as a regression problem to directly predict location coordinates and class probabilities for improving the detection speed, but with the precision reduced in general. However, with bags of efficient and effective tricks well used, one-stage detection methods, such as YOLOv4~\cite{bochkovskiy2020yolov4} can also achieve comparable or even better performance to two-stage detection methods. 

Because of the powerful feature extraction and representation ability, these CNN-based detection methods have been successfully applied to ship detection in SAR images. Based on two kinds of detection frameworks, effective network structures, training strategies and tricks are carefully designed to deal with multi-scale ship detection, leading to significant performance improvement. As for the two-stage SAR ship detectors, Jiao \textit{et al.}~\cite{jiao2018densely} fused different feature maps by a densely connected multi-scale neural network to solve multi-scale ship detection. Lin \textit{et al.}~\cite{lin2018squeeze} proposed a squeeze and excitation rank architecture to suppress redundant information of feature maps for representative ability improvement based on VGG network~\cite{simonyan2014very} pre-trained on ImageNet~\cite{Russakovsky2015ImageNet}. Cui \textit{et al.}~\cite{cui2019dense} adopted feature pyramid network (FPN)~\cite{lin2017feature} with convolutional block attention module (CBAM) densely connected to each feature map to integrate resolution and semantic information effectively. Wei \textit{et al.}~\cite{wei2020precise} introduced high-resolution ship detection network (HR-SDNet) to maintain high-resolution features, expecting to improve detection performance. Yan \textit{et al.}~\cite{zhao2020attention} proposed attention receptive pyramid network (ARPN) with receptive fields block (RFB) and convolutional block attention module (CBAM) combined reasonably. RFB can enhance local features with global dependency, while CBAM can boost useful information and suppress the influence of surroundings. As for the one-stage SAR ship detectors, Du \textit{et al.}~\cite{du2019saliency} effectively leverages saliency information to improve the representation capability of original SSD network~\cite{liu2016ssd} and enhance target detection results by focusing more on informative regions. Zhang \textit{et al.}~\cite{zhang2019depthwise} proposed a depth-wise separable convolution neural network that leverages multi-scale mechanism, concatenation mechanism and anchor box mechanism to improve ship detection speed dramatically. Inspired by the real-time idea of the YOLO algorithm, Zhang \textit{et al.}~\cite{zhang2019high} introduced a grid convolutional neural network equipped with depth-wise separable convolution to speed up the ship detection with little performance loss. Fu \textit{et al.}~\cite{fu2020anchor} introduced a feature balancing module for the small-scale ship detection and a feature-refinement module to tackle feature misalignment for better localization accuracies. Instead of horizontal bounding boxes, Chen \textit{et al.}~\cite{chen2020r2fa} and An \textit{et al.}~\cite{an2019drbox} used the oriented bounding boxes to perform more suitable ship detection for the geospatially arranged objects.
\begin{figure}[!htb]
	\centerline{\includegraphics[scale=0.3]{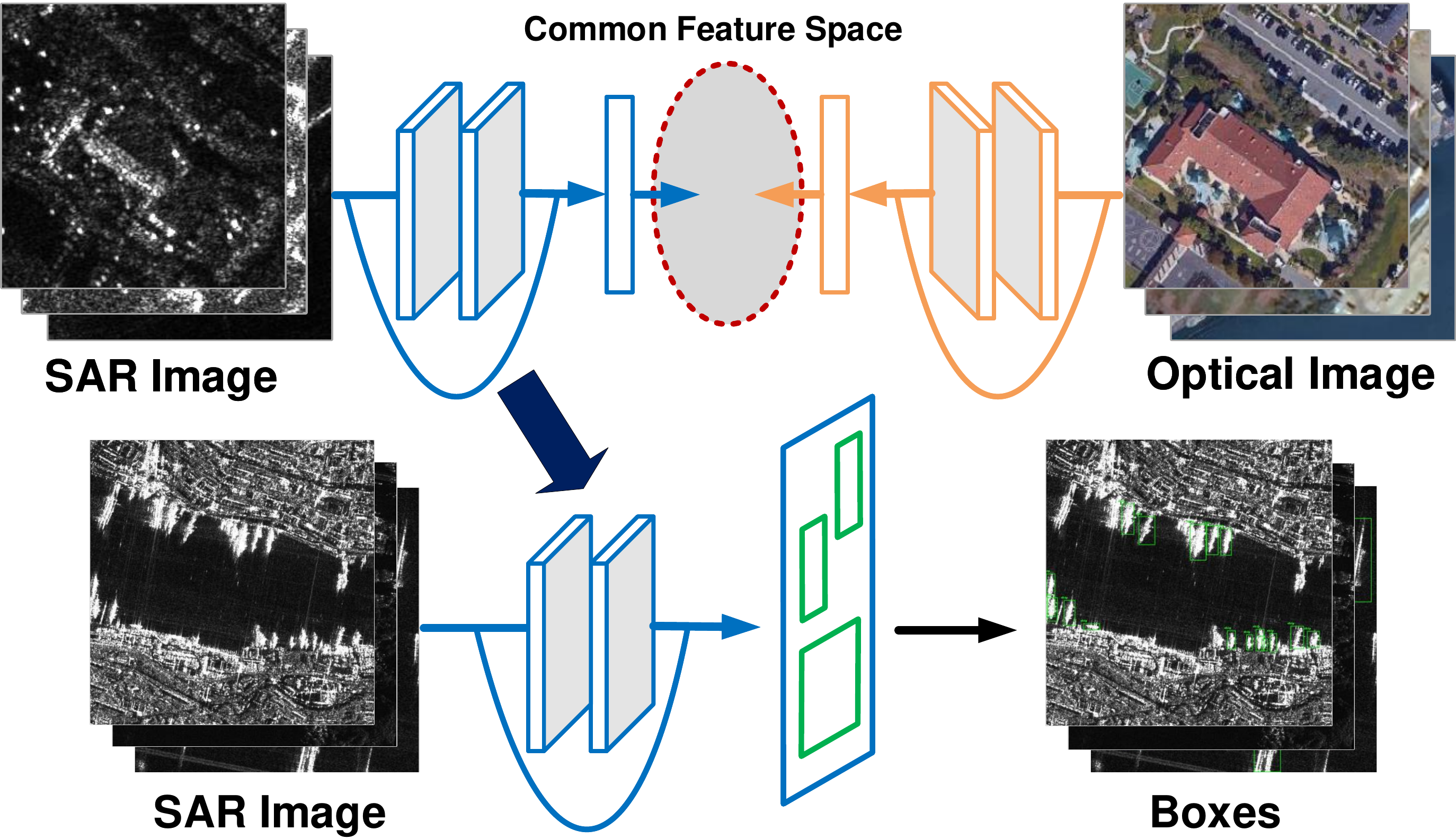}}
	\caption{Enhance the feature embedding of SAR ship detection via Optical-SAR common representation learning.}
	\label{fig:crl}
\end{figure}

On the one hand, the improvements brought by these detectors are attributed to effective network structures. On the other hand, the ImageNet~\cite{Russakovsky2015ImageNet} pretraining technique, a normal practice, is adopted to support all of these SAR ship detectors due to the scarcity of labeled SAR images. However, directly using ImageNet pretraining is hardly to obtain a good SAR ship detector, which is also a significant issue but with less attention paid. Specifically, one significant problem is the different imaging perspective. The ships in ImageNet are taken under natural scenes, while ships in SAR images are obtained from earth observations. The inconsistency across different ships' viewing perspective results in annotation information from ImageNet not applicable for SAR ship detectors. Thus, based on a large-scale aerial image dataset~\cite{xia2018dota}, we propose an optical ship detector (OSD) pretraining technique to improve feature learning ability of ships in SAR images. The OSD pretraining technique can transfer the characteristics of ships in earth observations to SAR images by fully taking advantage of a large amount of ships’ annotation information from aerial images. Another critical problem of directly applying ImageNet pretraining to SAR ship detection is the different imaging geometry between optical and SAR images, making ship detectors unable to obtain a powerful SAR feature embedding. Hence, we propose an Optical-SAR matching (OSM) pretraining technique to enhance the general feature embedding of SAR images. Specifically, the OSM pretraining technique transfers plentiful texture features from optical images to SAR images by common representation learning via bridge neural network (BNN)~\cite{xu2019task} on the optical-SAR matching task. As depicted in Fig~\ref{fig:crl}, BNN employs a couple of convolution neural networks (CNN) named left-CNN and right-CNN which project SAR images and optical images into a common feature space, respectively (see Section~\ref{section:bnn} for a more detailed discussion). The optical-SAR matching task forces BNN to learn useful fusion features. Then, the left-CNN can be further used as the backbone of the SAR detection framework to perform ship detection.

Based on the two pretraining techniques, we obtained the OSD pretraining based SAR ship detector (OSD-SSD) and the OSM pretraining based SAR ship detector (OSM-SSD), to solve the inconsistency across different imaging perspective and different imaging geometry, respectively. Furthermore, we propose to combine the two detectors to get a more comprehensively better detector based on the observation of their different advantages. Specifically, since the optical-SAR matching task mainly focuses on land area, the plentiful texture features from the optical images can help distinguish the building structures, resulting in fewer false alarms on land area using OSM-SSD. In contrast, the OSD-SSD can help identify and locate ships on sea area because of more ships' annotation information from the aerial image dataset. Considering these complementary advantages, we employ the weighted boxes fusion (WBF) strategy~\cite{solovyev2019weighted} to fuse the predictions of OSD-SSD and OSM-SSD. The WBF strategy utilizes confidence scores of all predicted bounding boxes to construct the averaged boxes including confidence scores and coordinate locations, leading to a SAR ship detector with better generalization ability.

\begin{figure}[!htb]
	\centerline{\includegraphics[scale=0.3]{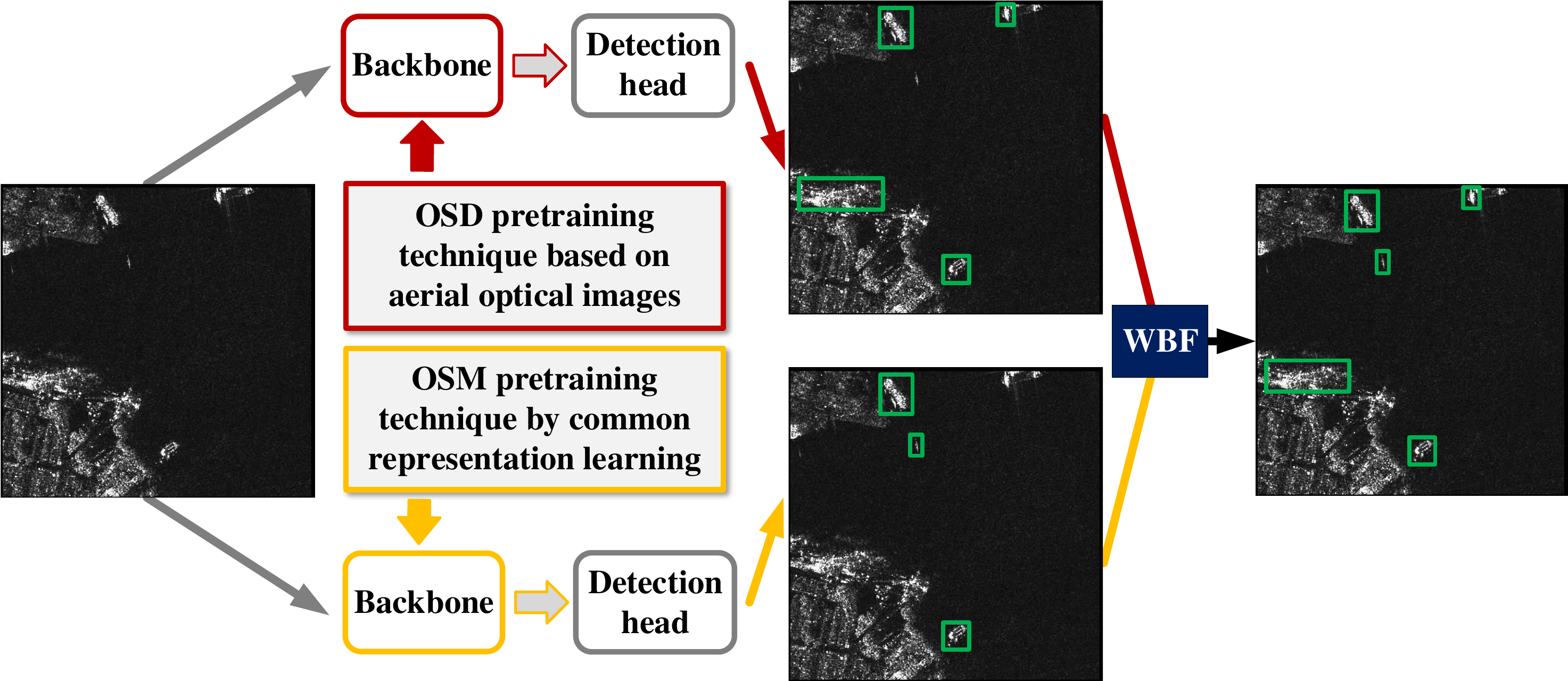}}
	\caption{Overall process to boost ship detection in SAR images.}
	\label{fig:over}
\end{figure}

According to the above analysis, instead of designing sophisticated network structures and using specific tricks, we consider the incompatibility between the ImageNet pretraining technique and SAR images, and propose two complementary pretraining techniques to boost SAR ship detection. The overall process is depicted in Fig.~\ref{fig:over}. Particularly, we utilize the OSD pretraining technique and OSM pretraining technique to obtain two complementary backbones, and then further train OSD-SSD and OSM-SSD, respectively. Finally, we leverage the WBF strategy to fuse the predictions of the two detectors. The main contributions of our work can be summarized into the following four aspects:

\begin{enumerate}
	\item {Improve the feature learning ability of ships in SAR images by proposing the OSD pretraining technique, which transfers the characteristics of ships in earth observations to SAR images by fully taking advantage of ships’ annotation information from a large-scale aerial image dataset~\cite{xia2018dota};}
	\item {Enhance the general feature embedding of SAR images by proposing the OSM pretraining technique, which transfers plentiful texture features from optical images to SAR images by common representation learning via bridge neural networks (BNN)~\cite{xu2019task} on the optical-SAR matching task;}
	\item {Explore the complementary characteristics of the OSD pretraining based SAR ship detector (OSD-SSD) and OSM pretraining based SAR ship detector (OSM-SSD) and thus propose to employ the WBF strategy~\cite{solovyev2019weighted} to fuse the predictions of the two detectors for further improving detection results;}
	\item {Conduct various experiments on four SAR ship detection datasets~\cite{xian2019air,competitation,wei2020hrsid,li2017ship} and two representative CNN-based detection benchmarks~\cite{ren2015faster,redmon2018yolov3} to verify the effectiveness and complementarity of the OSD-SSD and OSM-SSD detectors, and the state-of-the-art performance of the combination of the two detectors.}
\end{enumerate}
\begin{figure*}[!htb]
	\centering{\includegraphics[scale=0.6]{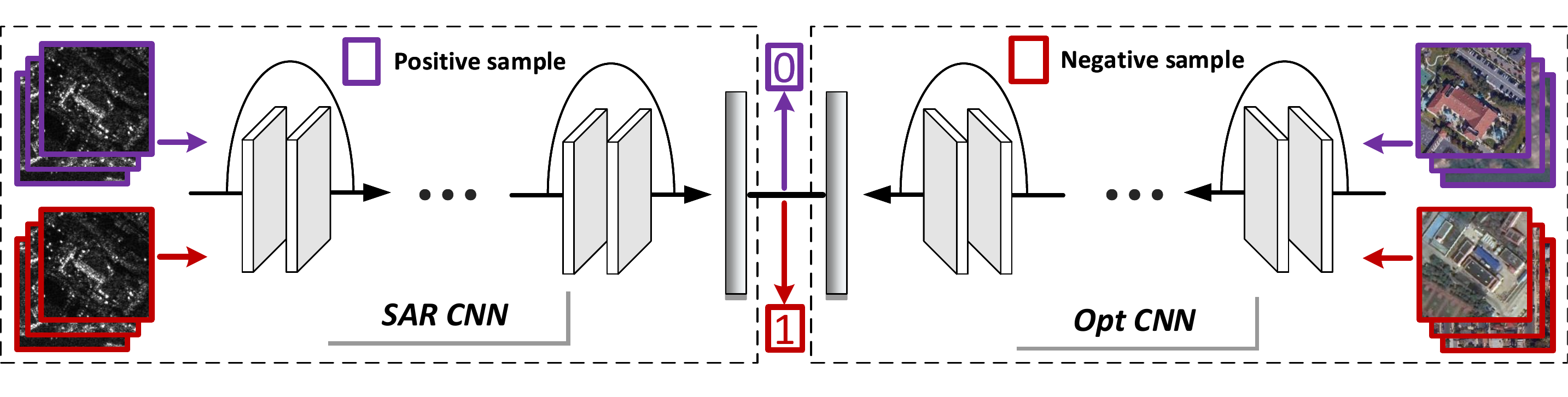}}
	\caption{Illustration of common representation learning via bridge neural network(BNN)~\cite{xu2019task} on the optical-SAR matching task, which adopts two convolutional neural networks named SAR CNN and Opt CNN to project SAR images and optical images into a common feature space. The Euclidean distance of the two output layers is regressed to 0 or 1 for positive samples or negative samples respectively.}
	\label{fig:bnn}
\end{figure*}

The rest of this paper is organized as follows. Section II introduces our methods in details. Section III provides the experimental settings and results analysis. Finally, some conclusions and future works are drawn in Section IV.

\section{METHODOLOGY}
In this section, we first introduce two CNN-based object detection benchmarks: Faster R-CNN~\cite{ren2015faster} and YOLOv3~\cite{redmon2018yolov3}. Next, the OSD pretraining technique based on aerial images and the OSM pretraining technique by common representation learning will be described in details. Finally, we will introduce how the WBF strategy~\cite{solovyev2019weighted} fuse the predictions of OSM-SSD and OSD-SSD.

\subsection{CNN-based object detection methods}
We select two representative methods: Faster R-CNN~\cite{ren2015faster} and YOLOv3~\cite{redmon2018yolov3}, as the benchmarks in our work. Here we only introduce the critical architecture of these two detection benchmarks, and we refer to their original paper~\cite{ren2015faster, redmon2018yolov3} to see a more detailed introduction. It is noted that our OSD pretraining technique, OSM pretraining technique and the WBF strategy~\cite{solovyev2019weighted} can be easily applied to other state-of-the-art object detection benchmarks.

\textbf{Faster R-CNN}~\cite{ren2015faster}, as the behalf of the two-stage object detection methods, consists of three modules: feature embedding network as the backbone to extract high-level features from the original images, region proposal network (RPN) generating the ship proposals for preliminarily predicting the location of ships and box regression network such as Fast R-CNN~\cite{girshick2015fast} finishing the binary classification and bounding box regression. We adopt ResNet50~\cite{he2016deep} as the backbone for better feature extraction. Besides, to handle the detection of multi-scale ships in multi-resolution SAR images, we use the feature pyramid network (FPN)~\cite{lin2017feature} to combine low-resolution, semantically strong features with high-resolution, semantically weak features. We initially set 3 anchor boxes with one scale of size 8 and three aspect ratios of size $\left\{0.5, 1, 2.0\right\}$ at each spatial location of each feature map. After RPN generates ship proposals, we adopt the RoIAlign~\cite{he2017mask} operation to fix the misalignment of feature maps caused by coarse spatial quantization. Furthermore, we select the cross entropy and $smooth$ L1 loss function to optimize the classification and regression task, respectively.

\textbf{YOLOv3}~\cite{redmon2018yolov3} is a representative one-stage object detection method comprising two modules: feature extraction network and box detection network. The feature extraction network is set as Darknet53~\cite{redmon2018yolov3} containing 53 convolutional layers and some shortcut connections as used in ResNet~\cite{he2016deep}, making YOLOv3 more powerful and efficient. As similarly used in Faster R-CNN, FPN is adopted in the feature extraction network to enhance multi-scale object detection capabilities. The box detection network predicts bounding boxes on three different scales following the feature extraction network. Feature maps with small sizes are used to detect large-size ships, while feature maps with large sizes are utilized to detect small-size ships. The predefined anchor boxes in Faster R-CNN are leveraged because learning the offsets between anchor boxes and the predictions will make the network easier to train. Considering this convenience, YOLOv3 also presets 9 anchors with 3 different sizes on each scale to predict bounding boxes more accurately. The anchor sizes are set according to the dataset and we use the same initial anchor sizes as~\cite{redmon2018yolov3}. Moreover, instead of fixing the input image size, we use the multi-scale training strategy to resize the input image into one size of
$\left\{320\times 320, 416\times 416\right\}$ randomly in each iteration, forcing the network to perform prediction well across different resolutions. Cross entropy and mean square error loss functions are utilized to optimize this network end-to-end.

\subsection{OSD pretraining technique based on aerial images}
The OSD pretraining technique transfers the characteristics of ships in earth observations to SAR images to improve the feature learning ability of ships in SAR images. Specifically, we first train optical ship detectors from aerial optical images and then leverage the backbone, ResNet50~\cite{he2016deep} for Faster R-CNN~\cite{ren2015faster} and Darknet53~\cite{redmon2018yolov3} for YOLOv3~\cite{redmon2018yolov3}, to further obtain OSD-SSD for fully taking advantage of ships' annotation information. As for the object detection in aerial images, recent advances have been witnessed because of the construction of many well-annotated datasets~\cite{cheng2016learning,xia2018dota} and efficient network design for specific problems~\cite{Ding_2019_CVPR}. 
Because a large number of high-resolution ships are available, we select DOTA~\cite{xia2018dota}, a large-scale dataset for object detection in aerial images, as the basic dataset to train our optical ship detectors. For the ResNet50 backbone, we directly extract it from the trained model provided by the RoI Transformer method~\cite{Ding_2019_CVPR}. Based on the Faster R-CNN benchmark, RoI Transformer applied a RRoI learner on region of interests (RoIs) to learn spatial transformation from horizontal proposals to oriented bounding box predictions, expecting to solve the common mismatch between horizontal box predictions and oriented objects. This transformation on feature maps enables the ResNet50 backbone more powerful feature extraction capabilities, especially for ships from a overlooking perspective. For extracting the Darknet53 backbone, we adopt the original YOLOv3 method to train the optical ship detector without any architectures modified. In addition to the ResNet50 and Darknet53 backbone, we can also extract FPN~\cite{lin2017feature} from the trained models using RoI Transformer and YOLOv3. We use ResNet50 (Darknet53) to denote ResNet50$+$FPN (Darknet53$+$FPN) for convenience.
\vspace{-0.1in} 
\subsection{OSM pretraining technique by common representation learning} 
\label{section:bnn}
The OSM pretraining technique transfers rich texture features from optical images to SAR images to obtain a specific feature extraction model with better SAR feature embedding capabilities. Specifically, the OSM pretraining technique resorts to a SAR feature embedding operator from common representation learning based on the optical-SAR matching task. As for common representation learning, several researches~\cite{eisenschtat2017linking,hughes2018identifying,xu2019task} were proposed to investigate the relationships between two data sources in different modalities. Among these methods, bridge neural network (BNN) first proposed in~\cite{xu2019task} is adopted to perform our common representation learning on the matching problem of optical and SAR images due to its excellent performance. As depicted in Fig~\ref{fig:crl}, BNN acts as a bridge and projects two images from different modalities into a common feature subspace. More specifically, given the optical-SAR matching task, we should first construct the data pairs of optical images and SAR images. Supposed we have two data sources denoted by $\left\{X_{s}, X_{o}\right\} \subset \mathbb{R}^{n_{1}} \times \mathbb{R}^{n_{2}}$, where $X_{s}=\left\{x^{i}_{s}\right\}^{N}_{i=1}$ is from SAR images and $X_{o}=\left\{x^{i}_{o}\right\}^{N}_{i=1}$ is from optical images. The i-th component $x^{i}_{s} \in X_{s}$, $x^{i}_{o} \in X_{o}$ are from the same region and match each other, while the i-th component $x^{i}_{s} \in X_{s}$, and the j-th component $x^{j}_{o} \in X_{o}$ are from different regions and do not match each other. Moreover, we define $D_{p}=\left\{x^{i}_{s}, x^{i}_{o}\right\}$ as positive samples and $D_{n}=\left\{x^{i}_{s}, x^{j}_{o}\right\}, i\neq j$ as negative samples. Secondly, instead of sharing weights, we build the BNN architecture illustrated in Figure~\ref{fig:bnn} which contains two separate, yet identical CNN: SAR CNN $f_{s}\left\{\cdot;\theta_{s}\right\}$ and Opt CNN $f_{o}\left\{\cdot;\theta_{2}\right\}$ with weights $\left(\theta_{s}, \theta_{o}\right)$.
For a pair of optical image and SAR image $\left(x_{s}, x_{o}\right)$, the output of SAR CNN and Opt CNN is $f_{s}\left(x_{s};\theta_{s}\right)$ and $f_{o}\left(x_{o};\theta_{o}\right)$, respectively. It is noted that the CNN can be replaced with ResNet50~\cite{he2016deep} and Darknet53~\cite{redmon2018yolov3} for the Faster R-CNN~\cite{ren2015faster} and  YOLOv3~\cite{redmon2018yolov3} benchmark, respectively. To decrease the feature dimension, following the backbone of the detection benchmark, we adopt a convolution layer with the filter size of $1 \times 1$, a batch normalization layer and a max-pooling layer to output an $m$-dimensional feature map. Finally, we add a linear layer followed by the sigmoid activation function to project the $m$-dimensional feature map into the common feature subspace. The BNN outputs the Euclidean distance of the two outputs of SAR CNN and Opt CNN, which is defined as follows:
\begin{equation}
f\left(x_{s}, x_{o} ; \theta_{s}, \theta_{o}\right)=\frac{1}{\sqrt{n}}\left\|\left(f_{s}\left(x_{s} ; \theta_{s}\right)-f_{o}\left(x_{o} ; \theta_{o}\right)\right)\right\|
\end{equation}
where $n$ is the dimension of the common feature. The ultimate goal of BNN is to determine whether the optical and SAR images have a potential relationship, i.e. positive sample or not. Thus the loss on positive sample set $D_{p}$ and negative sample set $D_{n}$ regresses the output $f\left(x_{s}, x_{o} ; \theta_{s}, \theta_{o}\right)$ to 0 if $\left(x_{s}, x_{o}\right) \in D_{P}$ and to 1 if $\left(x_{s}, x_{o}\right) \in D_{n}$ respectively, as follows:

\begin{equation}
l_{p}\left(D_{p} ; \theta_{s}, \theta_{o}\right)=\frac{1}{\left|D_{p}\right|} \sum_{\left(x_{s}, x_{o}\right) \in D_{p}}\left(f\left(x_{s}, x_{o} ; \theta_{s}, \theta_{o}\right)-0\right)^{2}
\end{equation}
\vspace{-0.1in} 
\begin{equation}
l_{n}\left(D_{n} ; \theta_{s}, \theta_{o}\right)=\frac{1}{\left|D_{n}\right|} \sum_{\left(x_{s}, x_{o}\right) \in D_{n}}\left(f\left(x_{s}, x_{o} ; \theta_{s}, \theta_{o}\right)-1\right)^{2}
\end{equation}

Thus, the problem of searching the common feature embeddings of optical and SAR images can be transferred to a binary classification problem where the overall loss of BNN on $D_{p}$ and $D_{n}$ can be designed as: 
\begin{equation}
L_{b n n}\left(D_{p}, D_{n} ; \theta_{s}, \theta_{o}\right)=\frac{l_{p}\left(D_{p} ; \theta_{s}, \theta_{o}\right)+\alpha \cdot l_{n}\left(D_{n} ; \theta_{s}, \theta_{o}\right)}{1+\alpha}
\end{equation}
where $\alpha$ is a hype-parameter to adjust the balance of positive
samples and negative samples. Then, the BNN model with the best weights $\left(\theta^{*}_{s}, \theta^{*}_{o}\right)$ can be find via
\begin{equation}
\left(\theta_{s}^{*}, \theta_{o}^{*}\right)=\operatorname{argmin}_{\theta_{s}, \theta_{o}} l\left(D_{p},D_{n} ; \theta_{s}, \theta_{o}\right)
\end{equation}
In the test phase, BNN uses a predefined threshold parameter $\gamma$ to decide whether the input data pair is positive sample or not. We train the BNN model on QXS-SAROPT dataset. Finally, we can exploit the SAR-CNN as the backbone of our OSM-SSD, expecting that the plentiful texture features from the optical image would enhance the feature embedding ability for SAR images.
\begin{figure}[!htb]
	\centerline{\includegraphics[scale=0.6]{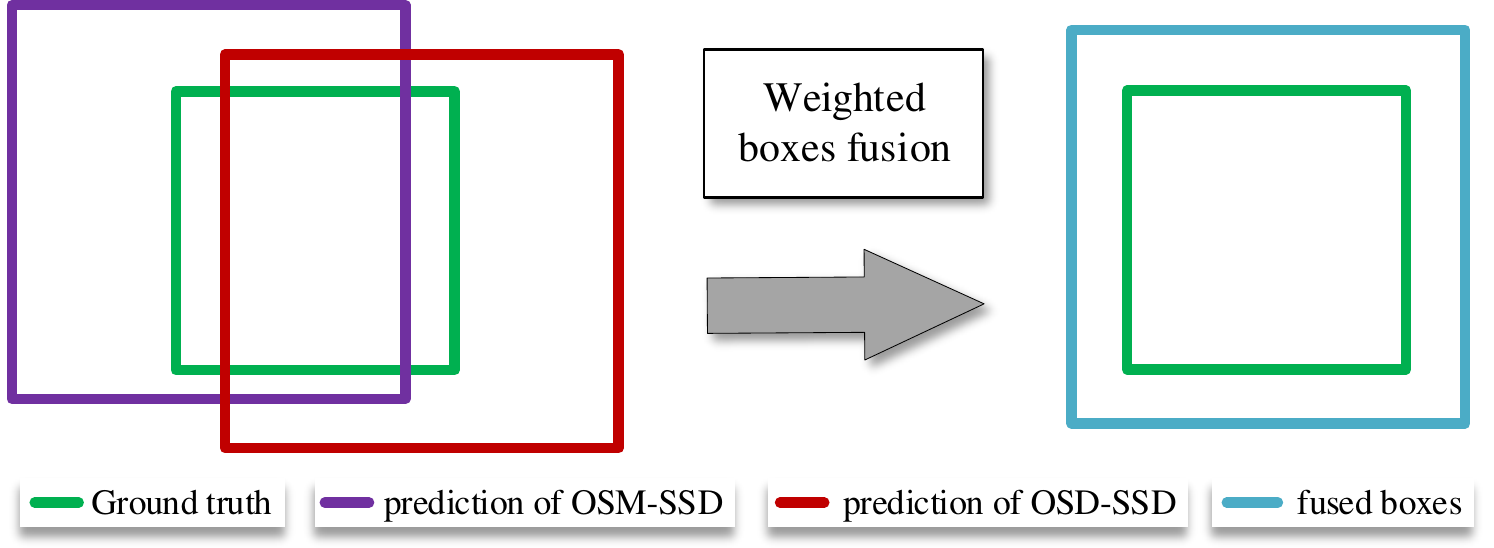}}
	\caption{Illustration of the WBF strategy~\cite{solovyev2019weighted}.}
	\label{fig:wbf}
\end{figure}

\begin{algorithm}[!htb]
	\caption{Weighted boxes fusion strategy. $B^{osd}$ and $B^{osm}$ are predicted boxes of OSD-SSD and OSM-SSD, respectively. $N$ is the number of models. $thr$ is the IoU threshold. Each predicted box is represented as $\left(Xmin, Ymin, Xmax, Ymax, C\right)$.}
	\begin{algorithmic}[1] 
		\vspace{0.1in} 
		\REQUIRE $N, B^{osd}, B^{osm}, thr$.
		\ENSURE Fused boxes of OSD-SSD and OSM-SSD.  
		\vspace{0.1in}         
		\STATE $B^{osd}$ and $B^{osm}$ are added to a list \textbf{L} and sorted in decreasing order according to the confidence score $C$ 
		\vspace{0.1in} 
		\STATE Declare an empty list \textbf{E} with each position storing a set of boxes (or a single box) for box clusters. Declare an empty list \textbf{F} with each corresponding position containing only one box to represent the fused box.
		\vspace{0.1in} 
		\STATE Loop through all predicted boxes in \textbf{L}, and attempt
		to find a matching box in the list \textbf{F}. If the IoU between a box in \textbf{F} and the current box in \textbf{L} is larger than $thr$, we define this box as a matching box ($IoU > thr$).
		\vspace{0.1in}
		\STATE If the matching box is not found in step $3$, add the current box from the list \textbf{L} to the end of lists \textbf{E} and \textbf{F} as new entries; proceed to the next box in the list \textbf{L}.
		\vspace{0.1in}
		\STATE If the matching box is found in step $3$, add the current box from the list \textbf{L} to \textbf{E} and the added position is the corresponding position of the matching box in \textbf{F}.
		\vspace{0.1in}
		\STATE Leveraging all $T$ boxes from the same position in \textbf{E} to recalculate the box coordinates and confidence score to form the fused box in \textbf{F} with the following equation:
		
		\begin{equation}
		C=\frac{\sum_{i=1}^{T} C_{i}}{T}
		\end{equation}
		
		\begin{equation}
		Xmin,max=\frac{\sum_{i=1}^{T} C_{i} \times Xmin_{i},max_{i}}{\sum_{i=1}^{T} C_{i}}
		\end{equation}
		
		\begin{equation}
		Ymin,max=\frac{\sum_{i=1}^{T} C_{i} \times Ymin_{i},max_{i}}{\sum_{i=1}^{T} C_{i}}
		\end{equation}
		
		where a box with larger confidence contributes more to the coordinates of the fused box than a box with lower confidence.
		\vspace{0.1in}
		\STATE After all boxes in the list \textbf{L} are traversed, adjust the confidence scores in \textbf{F} again with the following equation:
		\begin{equation}
		C=C \times \frac{T}{N}
		\end{equation}
	\end{algorithmic}
	\label{al:1}
\end{algorithm}

\subsection{Weighted boxes fusion}
We expect to leverage an effective fusion strategy to enhance detection performance based on the observation of the complementary advantages between OSD-SSD and OSM-SSD. A commonly used fusion strategy is non maximum suppression (NMS). NMS sorts all the predicted bounding boxes according to classification confidence and removes redundant boxes to keep the one with the highest confidence score. However, NMS pays too much attention to classification confidence without considering localization accuracy. For example, if a predicted box has a high intersection over union (IoU) with ground truth but a low confidence score, it will be removed by another predicted box with high confidence score but a low IoU. Although soft-NMS~\cite{bodla2017soft} can alleviate this problem, it will retain largely redundant boxes, increasing many false alarms. In order to simultaneously take classification confidence and localization accuracy into account, we adopt the weighted boxes fusion(WBF) strategy~\cite{solovyev2019weighted} to combine the predictions of OSM-SSD and OSD-SSD, expecting to improve the generalization ability of SAR ship detectors. The WBF strategy in our method utilizes confidence scores of predicted bounding boxes to construct the averaged boxes, including not only averaged confidence scores but also averaged localization predictions as depicted in Fig~\ref{fig:wbf}. More specifically, a predicted box with higher classification confidence will more proportionally contribute to the final averaged box. The process of WBF strategy is illustrated in Alg~\ref{al:1}.

\section{Experiments}
\subsection{Datasets}
We conduct experiments on three kinds of datasets: DOTA used for OSD pretraining, QXS-SAROPT used for OSM pretraining and four SAR ship detection datasets used for OSD-SSD and OSM-SSD. It is noted that the OSD-pretrained model based on the Faster R-CNN~\cite{ren2015faster} benchmark can be directly obtained from~\cite{Ding_2019_CVPR}, so we only perform OSD pretraining using the YOLOv3~\cite{redmon2018yolov3} benchmark on DOTA. 

\textbf{DOTA}~\cite{xia2018dota} is a large-scale dataset for object detection in aerial images including 2806 aerial images with each image ranging from $800 \times 800$ to $4000 \times 4000$ pixels collected from different sensors and platforms. Compared to version 1.0 with 15 categories annotated, the fully annotated DOTA version 1.5 (DOTA-v1.5) contains 16 common object categories, including the ship. We used DOTA-v1.5 and adopted the same dataset split strategy as used in~\cite{Ding_2019_CVPR}: half of the original images are selected as the training set, 1/6 as the validation set, and 1/3 as the testing set. We cropped each image into $1024 \times 1024$ pixels with the overlap of 256 pixels and trained the optical ship detectors on the training set and directly report the performance of detectors on the validation set because the annotation information of test set is not available.

\textbf{QXS-SAROPT} includes 20,000 pairs of corresponding SAR and optical images extracted from GaoFen-3  high-resolution spotlight images and Google Earth remote sensing optical images. The size of each image is $256 \times 256$. We select 14,000 image pairs as the training set and the remaining 6,000 image pairs as the test set to train our BNN.

\textbf{SAR ship detection datasets}. We conduct SAR ship detection experiments on four datasets: AIR-SARShip-1.0~\cite{xian2019air}, AIR-SARShip-2.0~\cite{competitation} which is also the Gaofen competition dataset provided by the ``2020 Gaofen Challenge on automated high-resolution earth observation image interpretation'', HRSID~\cite{wei2020hrsid} and SSDD~\cite{li2017ship}. The AIR-SARShip-1.0 dataset comprises 31 high-resolution large-scale $3000 \times 3000$ images collected from the GF-3 satellite. We randomly select 21 images as the training and validation data, and the remaining 10 images as the test data. The AIR-SARShip-2.0 dataset includes 300 images of size $1000 \times 1000$ with spatial resolution ranging from 1m to 5m collected from the Gaofen-3 satellite. We randomly select 210 images as training and validation data, and the remaining 90 images are used for test data. Each image of both AIR-SARShip-1.0 and AIR-SARShip-2.0 is cropped into $512 \times 512$ pixels with an overlap of 256 pixels for training and testing. The HRSID dataset contains 5604 cropped SAR images with a size of $800 \times 800$ and has been divided into a training set and a test set at a ratio of 65 to 35. The SSDD dataset contains 1160 images of resolution from 1m to 15 m in total, where the training set includes 928 images and the test set includes the remaining 232 images. For the Faster R-CNN benchmark, we directly input each image of AIR-SARShip-1.0, AIR-SARShip-2.0 and HRSID into the network, and resize each image of SSDD into $800 \times 800$ pixels. As for the YOLOv3 benchmark, all the images of the four datasets are resized into $416 \times 416$ pixels as the input of the detector. We do not apply any data augmentation method for all datasets except for the scaling technique because of the multi-scaling training strategy for the YOLOv3 benchmark.
\subsection{Parameter settings}
All the experiments are implemented in the PyTorch 1.7 framework and carried out over an NVIDIA 3070 GPU. The PC operating system is a 64-bit Ubuntu 20.04. 
\subsubsection{OSD pretraining technique} For the Faster R-CNN~\cite{ren2015faster} benchmark, we directly extract the ResNet50~\cite{he2016deep} backbone from the existing trained model from~\cite{Ding_2019_CVPR}. As for the YOLOv3~\cite{redmon2018yolov3} benchmark, we first use MMDetection (https://github.com/open-mmlab/mm-detection) to train a optical ship detector and then extract the Darknet53~\cite{redmon2018yolov3} backbone from the trained detector. The optical ship detector is trained with stochastic gradient descent (SGD) for 240 epochs with a total of 12 images per minibatch. The initial learning rate is set as 0.001, which is then divided by a factor of 10 at the 160th and 200th epoch. The weight decay is 0.0005 and the momentum is 0.9. The ImageNet based pre-trained ResNet50 and Darknet53 are utilized for a better converge point.
\subsubsection{OSM pretraining technique} Our BNN model based on ResNet50~\cite{he2016deep} and Darknet53~\cite{redmon2018yolov3} are both trained with SGD for 200 epochs with a batch size of 20. The initial learning rate is set as 0.01 and then divided by a factor of 2 at the 30th and 100th epochs. The dimension of the common feature is set as 50 and the adjusting factor $\alpha$ is 1. The threshold $\gamma$ is set as 0.5. We also adopt ImageNet based pre-trained model to train BNN.
\subsubsection{SAR ship detectors} We also use MMDetection to implement OSD-SSD and OSM-SSD. For the Faster R-CNN~\cite{ren2015faster} benchmark, all models are trained with SGD for 14 epochs with a total of eight images per minibatch. The initial learning rate is set as 0.02, and then divided by a factor of 10 at the 8th and 12th epochs. The weight decay is 0.0001 and the momentum is 0.9. For the YOLOv3~\cite{redmon2018yolov3} benchmark, all models are trained with SGD for 240 epochs with a total of 12 images per minibatch. The initial learning rate is set as 0.001, which is then divided by a factor of 10 at the 160th and 200th epochs. The IoU threshold is set as 0.5 when training and testing for rigorous filtering the bounding boxes with low precision. Warm-up~\cite{he2016deep} is introduced during the initial training stage to avoid gradient explosion and the corresponding number of iterations is 500. We use the same settings for all experiments for a fair comparison.
\subsubsection{WBF} We fuse predicted boxes from OSD-SSD and OSM-SSD through the WBF strategy~\cite{solovyev2019weighted}. The IoU threshold $thr$ is set as 0.7 verified by many experiments and see Section~\ref{section:wbf} for a detailed discussion. 
\subsection{Evaluation Metrics}
Precision, recall and F1 scores are employed to evaluate the performance of SAR ship detectors, and the definition of these evaluation metrics is given as follows: 
\begin{equation}
\rm Precision=\frac{N_{TP}}{N_{TP}+N_{FP}}
\end{equation}
\begin{equation}
\rm Recall=\frac{N_{TP}}{N_{TP}+N_{FN}}
\end{equation}
\begin{equation}
\rm F1=\frac{2 \times Precision \times  Recall}{Precision+Recall}
\end{equation}
where $TP$ is True Positive, $FP$ is False Positive, $TN$ is True Negative and $FN$ is False Negative. $N_{TP}$, $N_{FP}$, $N_{FN}$ is the number of $TP$, $FP$ and $FN$, respectively. More specifically, $TP$ indicates the correctly detected ships, $FP$ represents the false alarms and $FN$ denotes the missing ships. A predicted bounding box is considered as a true positive if its IoU with the ground truth is higher than a given IoU threshold i.e. 0.5. Otherwise, it is regarded as a false positive. Moreover, the predicted bounding boxes with the highest confidence score are seen as the true positive, if the IoU of several ones with the ground truth are all higher than the threshold. F1 score is a comprehensive evaluation metric for the quantitative performance of different models by simultaneously considering the precision rate and recall rate. To further evaluate the comprehensive quality of SAR ship detectors, we also adopt the average precision (AP) metric, which can be defined as:
\begin{equation}
\mathrm{AP}=\frac{1}{101} \sum_{r \in \mathrm{S}} \text { Precision }\left.\right|_{\text {Recall }=r}
\end{equation}
where $S=\left\{0, 0.01, ..., 1\right\}$ representing a set of equally spaced
recall rates. $\rm AP_{0.5}$ denote the IoU threshold being 0.5. $\rm AP$ indicates that the IoU threshold is set from 0.50 to 0.95 with the step size set as 0.05.
\subsection{Results Analysis}
\vspace{0.1in}
\renewcommand{\arraystretch}{1.1} 
\begin{table}[!htb]  
	\centering
	\setlength{\tabcolsep}{1mm}   
	\begin{threeparttable} 
		\caption{Detection results of optical ship detectors on the DOTA dataset~\cite{xia2018dota}.}
		\begin{tabular}{ccccccc}
			\toprule
			\multirow{2}{*}{Benchmark}
			&\multicolumn{3}{c}{Ship}&\multicolumn{3}{c}{All categories} \\
			\cmidrule(lr){2-4} \cmidrule(lr){5-7}
			&${\rm AP_{0.5}}$&${\rm AP_{0.75}}$&${\rm AP}$&${\rm mAP_{0.5}}$&${\rm mAP_{0.75}}$&${\rm mAP}$ \\
			\midrule  
			Faster R-CNN~\cite{ren2015faster}&$0.807$&$-$&$-$&$0.650$&$-$&$-$ \cr 
			YOLOv3~\cite{redmon2018yolov3}&$0.605$&$0.477$&$0.470$&$0.509$&$0.284$&$0.287$ \cr
			\bottomrule
		\end{tabular} 
		\label{tab:dota} 
	\end{threeparttable}  
\end{table}
\renewcommand{\arraystretch}{1.2} 
\begin{table}[!htb]  
	\centering
	\setlength{\tabcolsep}{1mm}   
	\begin{threeparttable} 
		\caption{Overall performance of different methods on four SAR ship detection datasets using the Faster R-CNN~\cite{ren2015faster} benchmark.}
		\begin{tabular}{ccccccc}
			\toprule
			\multicolumn{7}{c}{\textbf{AIR-SARShip-1.0~\cite{xian2019air}}} \\
			\midrule
			Method&${\rm Precision}$&${\rm Recall}$&$\rm F1$&${\rm AP_{0.5}}$&${\rm AP_{0.75}}$&${\rm AP}$ \\
			\midrule  
			ImageNet-SSD&$0.9232$&$0.8003$&$0.8574$&$0.8720$&$0.5461$&$0.5129$\cr 
			OSD-SSD&$\mathbf{0.9436}$&$0.8529$&$\mathbf{0.8960}$&$0.8921$&$0.6260$&$0.5605$\cr
			OSM-SSD&$0.9324$&$0.8248$&$0.8753$&$0.8852$&$0.6153$&$0.5586$\cr
			WBF-DM&$0.9045$&$\mathbf{0.8653}$&$0.8847$&$\mathbf{0.9006}$&$\mathbf{0.6452}$&$\mathbf{0.5801}$\cr
			\midrule  
			\multicolumn{7}{c}{\textbf{AIR-SARShip-2.0~\cite{competitation}}} \\
			\midrule
			ImageNet-SSD&$0.8561$&$0.7964$&$0.8252$&$0.8487$&$0.5633$&$0.5190$\cr 
			OSD-SSD&$0.8563$&$0.8232$&$0.8391$&$0.8626$&$0.6028$&$0.5618$\cr
			OSM-SSD&$\mathbf{0.8676}$&$0.8170$&$0.8415$&$0.8582$&$0.6129$&$0.5528$\cr
			WBF-DM&$0.7944$&$\mathbf{0.8482}$&$\mathbf{0.8677}$&$\mathbf{0.8736}$&$\mathbf{0.6451}$&$\mathbf{0.5880}$\cr
			\midrule  
			\multicolumn{7}{c}{\textbf{HRSID~\cite{wei2020hrsid}}} \\
			\midrule
			ImageNet-SSD&$0.8842$&$0.8624$&$0.8736$&$0.8878$&$0.7851$&$0.6703$\cr 
			OSD-SSD&$\mathbf{0.8975}$&$0.8617$&$\mathbf{0.8784}$&$0.8892$&$0.7851$&$0.6743$\cr
			OSM-SSD&$0.8826$&$0.8658$&$0.8751$&$0.8932$&$0.7890$&$0.6719$\cr
			WBF-DM&$0.8455$&$\mathbf{0.8823}$&$0.8736$&$\mathbf{0.8971}$&$\mathbf{0.7988}$&$\mathbf{0.6844}$\cr
			\midrule  
			\multicolumn{7}{c}{\textbf{SSDD~\cite{li2017ship}}} \\
			\midrule
			ImageNet-SSD&$0.9332$&$0.9440$&$0.9385$&$0.9615$&$0.7091$&$0.6124$\cr 
			OSD-SSD&$0.9417$&$0.9558$&$\mathbf{0.9487}$&$0.9704$&$0.7485$&$0.6328$\cr
			OSM-SSD&$\mathbf{0.9429}$&$0.9485$&$0.9457$&$0.9679$&$0.7427$&$0.6228$\cr
			WBF-DM&$0.9079$&$\mathbf{0.9669}$&$0.9364$&$\mathbf{0.9740}$&$\mathbf{0.7534}$&$\mathbf{0.6426}$\cr
			\bottomrule
		\end{tabular} 
		\label{tab:ap1} 
	\end{threeparttable}  
\end{table}
\renewcommand{\arraystretch}{1.2} 
\begin{table}[!htb]  
	\centering
	\setlength{\tabcolsep}{1mm}   
	\begin{threeparttable} 
		\caption{Overall performance of different methods on four SAR ship detection datasets using the YOLOv3~\cite{redmon2018yolov3} benchmark.}
		\begin{tabular}{ccccccc}
			\toprule
			\multicolumn{7}{c}{\textbf{AIR-SARShip-1.0~\cite{xian2019air}}} \\
			\midrule
			Method&${\rm Precision}$&${\rm Recall}$&$\rm F1$&${\rm AP_{0.5}}$&${\rm AP_{0.75}}$&${\rm AP}$ \\
			\midrule  
			ImageNet-SSD&$0.8901$&$0.8603$&$0.8744$&$0.8712$&$0.5546$&$0.5024$\cr 
			OSD-SSD&$\mathbf{0.9411}$&$0.8740$&$\mathbf{0.9062}$&$0.8849$&$0.6273$&$0.5398$\cr
			OSM-SSD&$0.9324$&$0.8797$&$0.9055$&$0.8836$&$0.6146$&$0.5375$\cr
			WBF-DM&$0.8667$&$\mathbf{0.8964}$&$0.8808$&$\mathbf{0.8913}$&$\mathbf{0.6529}$&$\mathbf{0.5672}$\cr
			\midrule  
			\multicolumn{7}{c}{\textbf{AIR-SARShip-2.0~\cite{competitation}}} \\
			\midrule
			ImageNet-SSD&$0.8711$&$0.7960$&$0.8323$&$0.8300$&$0.4622$&$0.4649$\cr 
			OSD-SSD&$0.9064$&$0.8172$&$\mathbf{0.8593}$&$0.8470$&$0.5609$&$0.5096$\cr
			OSM-SSD&$\mathbf{0.9091}$&$0.8150$&$0.8562$&$0.8449$&$0.5509$&$0.5114$\cr
			WBF-DM&$0.8584$&$\mathbf{0.8497}$&$0.8535$&$\mathbf{0.8543}$&$\mathbf{0.6156}$&$\mathbf{0.5478}$\cr
			\midrule  
			\multicolumn{7}{c}{\textbf{HRSID~\cite{wei2020hrsid}}} \\
			\midrule
			ImageNet-SSD&$0.8541$&$0.8593$&$0.8567$&$0.8364$&$0.5538$&$0.5229$\cr 
			OSD-SSD&$\mathbf{0.9010}$&$0.8684$&$\mathbf{0.8841}$&$0.8512$&$0.6035$&$0.5496$\cr
			OSM-SSD&$0.8662$&$0.8658$&$0.8665$&$0.8454$&$0.5731$&$0.5319$\cr
			WBF-DM&$0.7879$&$\mathbf{0.8933}$&$0.8375$&$\mathbf{0.8796}$&$\mathbf{0.6581}$&$\mathbf{0.5850}$\cr
			\midrule  
			\multicolumn{7}{c}{\textbf{SSDD~\cite{li2017ship}}} \\
			\midrule
			ImageNet-SSD&$0.9482$&$0.9509$&$0.9493$&$0.9447$&$0.6361$&$0.5840$\cr 
			OSD-SSD&$\mathbf{0.9614}$&$0.9505$&$0.9556$&$0.9474$&$0.6572$&$0.5898$\cr
			OSM-SSD&$0.9571$&$0.9560$&$\mathbf{0.9561}$&$0.9468$&$0.6759$&$0.5885$\cr
			WBF-DM&$0.9413$&$\mathbf{0.9633}$&$0.9523$&$\mathbf{0.9577}$&$\mathbf{0.7017}$&$\mathbf{0.6058}$\cr
			\bottomrule
		\end{tabular} 
		\label{tab:ap2} 
	\end{threeparttable}  
\end{table}
\vspace{0.2in}
\subsubsection{OSD-SSD} Table~\ref{tab:dota} shows the detection results of our optical ship detectors on DOTA dataset~\cite{xia2018dota} including the mean AP (mAP) for all categories and $\rm AP$ for the ship category. It is noted that the results of the Faster R-CNN~\cite{ren2015faster} benchmark is directly from~\cite{Ding_2019_CVPR} ('-' means that the original paper didn't provide the corresponding results). We can see that $\rm AP_{0.5}$ achieves $80.7\%$ and $60.5\%$ for the Faster R-CNN and YOLOv3~\cite{redmon2018yolov3} benchmark respectively, indicting that ResNet50~\cite{he2016deep} and Darknet53~\cite{redmon2018yolov3} backbone both have excellent ship detection capabilities. As for SAR ship detectors, the detection performance is showed in Table~\ref{tab:ap1} and Table~\ref{tab:ap2} for the Faster R-CNN and YOLOv3 benchmark, respectively. It is observed that OSD-SSD outperforms the ImageNet pretraining based SAR ship detector (ImageNet-SSD) under different metrics. Take the detection results on AIR-SARShip-2.0~\cite{competitation} using the YOLOv3 benchmark for an example, the precision rate of the OSD-SSD model gains a large improvement of 3.53\% and the recall rate achieves 2.12\% higher value. Due to these improvements, the OSD-SSD model finally achieves 2.7\% higher F1 score and 1.70\% higher $\rm AP_{0.5}$, reflecting overall performance improvements. Furthermore, when the IoU threshold becomes larger indicting the requirement of localization accuracy gets higher, $\rm AP_{0.75}$ and $\rm AP$ gain a larger improvement of $9.87\%$ and $4.47\%$ respectively, which means that the predicted bounding boxes are more accurate. Similar phenomena are also present on other datasets for both the Faster R-CNN and YOLOv3 benchmark, demonstrating the superiority of our OSD-SSD model. In other words, compared to ships in the natural scene from ImageNet, ships' annotation information from earth observations can better help improve the feature learning ability of ships in SAR images.
\renewcommand{\arraystretch}{1.1} 
\begin{table}[!htb]  
	\centering
	\setlength{\tabcolsep}{4mm}   
	\begin{threeparttable} 
		\caption{Matching results of BNN~\cite{xu2019task} on the QXS-SAROPT dataset.}
		\begin{tabular}{cccc}
			\toprule
			Backbone&Accuracy&Precision&Recall \\
			\midrule  
			ResNet50~\cite{he2016deep}&$0.829$&$0.748$&$0.993$ \cr 
			Darknet53~\cite{redmon2018yolov3}&$0.828$&$0.746$&$0.995$ \cr
			\bottomrule
		\end{tabular} 
		\label{tab:bnn} 
	\end{threeparttable}  
\end{table}
\renewcommand{\arraystretch}{1.3} 
\begin{table*}[!htb]  
	\centering
	\setlength{\tabcolsep}{4mm}   
	\begin{threeparttable} 
		\caption{Performance of ImageNet-SSD, OSD-SSD and OSM-SSD under two different scenes.}
		\begin{tabular}{c|c|cccccc}
			\hline
			\multicolumn{1}{c|}{Scene}&\multicolumn{1}{c|}{Method}
			&Precision&Recall&F1&$\rm AP_{0.5}$&$\rm AP_{0.75}$&$\rm AP$ \\
			\hline
			\multirow{3}{*}{Inshore}
			&ImageNet-SSD&$0.8484$&$0.8139$&$0.8307$&$0.8777$&$0.5336$&$0.4941$ \cr
			&OSD-SSD&$0.8655$&$\mathbf{0.8604}$&$\mathbf{0.8629}$&$\mathbf{0.8973}$&$\mathbf{0.5982}$&$\mathbf{0.5332}$ \cr
			&OSM-SSD&$\mathbf{0.8805}$&$0.8197$&$0.8491$&$0.8876$&$0.5789$&$0.5112$ \cr
			\hline  
			\multirow{3}{*}{Offshore}
			&ImageNet-SSD&$0.9640$&$0.9919$&$0.9777$&$0.9879$&$0.7861$&$0.6605$ \cr
			&OSD-SSD&$0.9634$&$\mathbf{0.9946}$&$\mathbf{0.9788}$&$0.9881$&$\mathbf{0.8203}$&$\mathbf{0.6705}$ \cr 
			&OSM-SSD&$\mathbf{0.9659}$&$0.9919$&$0.9787$&$\mathbf{0.9885}$&$0.8074$&$0.6663$ \cr
			\hline
		\end{tabular} 
		\label{tab:in-off} 
	\end{threeparttable}  
\end{table*}
\subsubsection{OSM-SSD}
Table~\ref{tab:bnn} suggests that our BNN has an outstanding performance on the QXS-SAROPT dataset for both ResNet50~\cite{he2016deep} and Darknet53~\cite{redmon2018yolov3} backbone. Specifically, the pair matching accuracy based on ResNet50 and Darknet53 are $82.9\%$ and $82.8\%$ , respectively, demonstrating that BNN can well predict the relationship of SAR and optical images, and obtain useful common features. As for the performance of OSM-SSD based on the Faster R-CNN~\cite{ren2015faster} and YOLOv3~\cite{redmon2018yolov3} benchmark, the overall results under different metrics are also illustrated in Table~\ref{tab:ap1} and Table~\ref{tab:ap2}, respectively. We can clearly see that, our OSM-SSD performs better than ImageNet-SSD under all evaluation metrics for different network architectures and different datasets. Especially on AIR-SARShip-2.0~\cite{competitation} using the YOLOv3 network, 3.80\%, 1.90\%, 2.39\%, 1.49\%, 8.87\% and 4.65\% performance improvement can be achieved in terms of precision rate, recall rate, F1 score, $\rm AP_{0.5}$, $\rm AP_{0.75}$ and $\rm AP$, respectively. All these improved performances can prove that the common features obtained from common representation learning can boost ship detection in SAR images. In other words, based on the optical-SAR matching task, BNN can well transfer rich texture features from aerial optical images to SAR images, enhancing the feature extraction capability of SAR ship detectors without additional ships' annotation information and any network architecture modified. 

\vspace{0.1in}
\begin{figure*}[!htb]
	\centering{\includegraphics[scale=0.64]{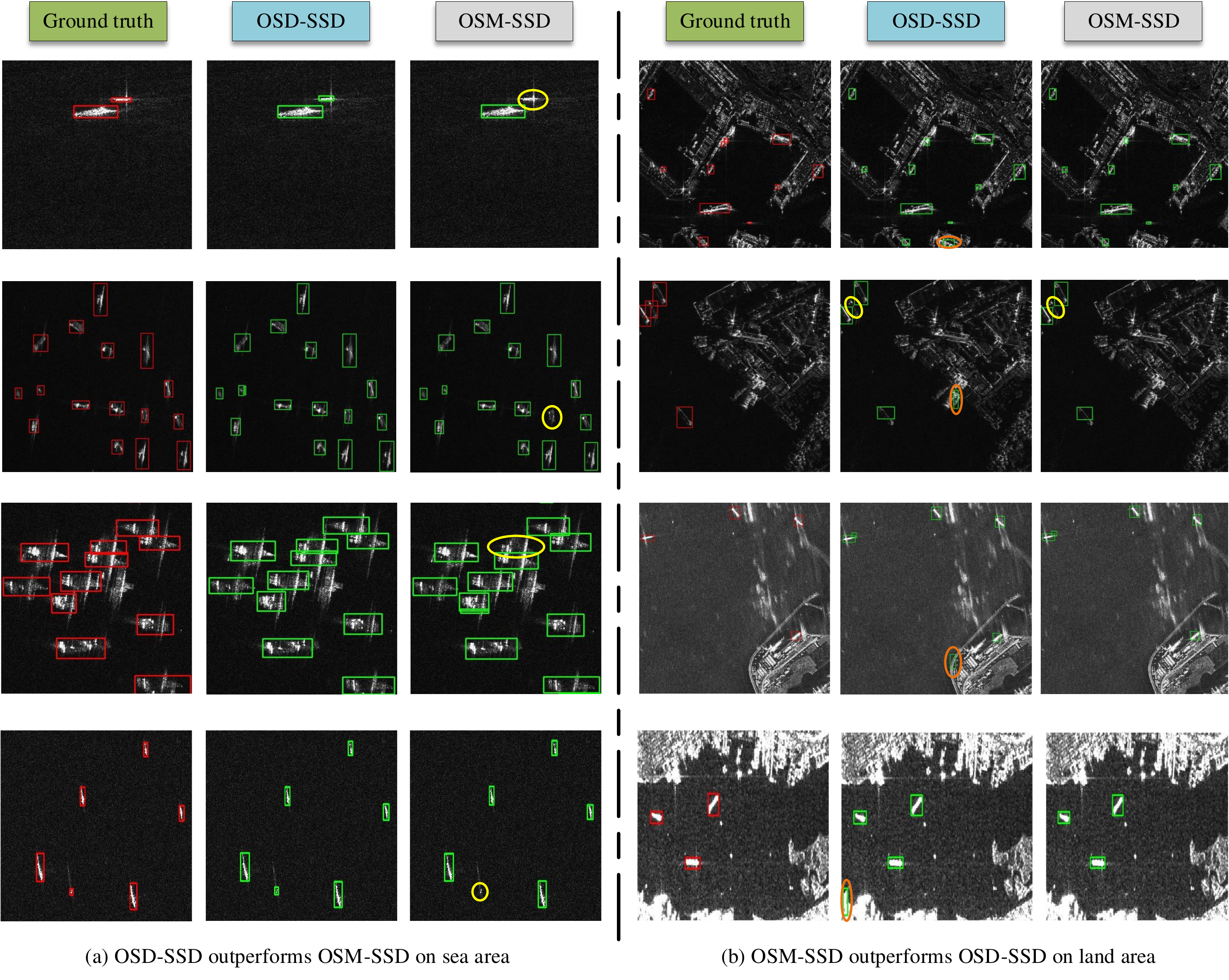}}
	\caption{Comparison results of OSD-SSD and OSM-SSD on four SAR ship detection datasets. The red rectangles are ground truth and the green rectangles represent the detection results. The orange and yellow circles denote the false alarms and missing ships, respectively. Two sets of visualization results demonstrate the complementary characteristics of OSD-SSD and OSM-SSD: (a) OSD-SSD outperforms OSM-SSD on sea area. (b) OSM-SSD outperforms OSD-SSD on land area.}
	\label{fig:vis}
\end{figure*}

\subsubsection{Comparison between OSM-SSD and OSD-SSD}
As listed on AIR-SARShip2.0 using the Faster R-CNN~\cite{ren2015faster} benchmark from Table~\ref{tab:ap1}, OSD-SSD and OSM-SSD achieve comparable detection results in different degrees, such as 1.39\%, 3.95\%, 4.28\% improvements verse 0.95\%, 4.96\%, 3.38\% improvements in terms of $\rm AP_{0.5}$, $\rm AP_{0.75}$ and $\rm AP$, respectively. It is noted that the DOTA dataset~\cite{xia2018dota} on which our OSD pretraining performed has a large amount of annotation information, while the QXS-SAROPT dataset on which our OSM pretraining performed only has region matching information without any ship annotation. Moreover, we also conduct experiments under the inshore and offshore scenes, respectively. It can be seen from Table~\ref{tab:in-off} that OSD-SSD and OSM-SSD have different performances in different scenes. Specifically, compared to OSD-SSD, our OSM-SSD gains 1.50\% improvements in precision rate under inshore scenes. As for the offshore ship detection, our OSD-SSD improves the recall rate by 0.25\% than OSM-SSD. In addition to quantitative comparisons, we also visualize some detection results in Fig.~\ref{fig:vis} to show an intuitive understanding of their effect. We use red rectangles, green rectangles, orange circles and yellow circles to figure out ground truth, predicted boxes, false alarms and missing ships, respectively. In Fig.~\ref{fig:vis} (a), considering the first scene on sea area, two ground truth are perfectly detected using OSD-SSD while one ship is missed in the prediction of OSM-SSD. Similar phenomenon that OSD-SSD has a better recall on sea area also occurs in other scenes. We conjecture that this observation is because the OSD-pretrained model can better capture ship features for large amounts of ship annotation in the aerial optical dataset. On the contrary, we can see that OSM-SSD has a better performance than OSD-SSD on the land area from Fig.~\ref{fig:vis} (b). Taking the first row as an example, compared to one false alarm on land area for OSD-SSD, OSM-SSD accurately predicts the ground truth without any false alarms. What cause this phenomenon is that the optical-SAR matching dataset on which our OSM pretraining performed mainly focuses on land area, leading to OSM-SSD has better feature embeddings and fewer false alarms on land area. In conclusion, OSD-SSD and OSM-SSD can improve detection results from different perspectives, proving the effectiveness of our OSD and OSM pretraining techniques. More importantly, these complementary advantages between OSD-SSD and OSM-SSD straightly inspire us to utilize fusion strategies to boost ship detection results, which will be analyzed in the next part. 

\subsubsection{WBF-DM}
\begin{figure}[!htb]
	\centering{\includegraphics[scale=0.6]{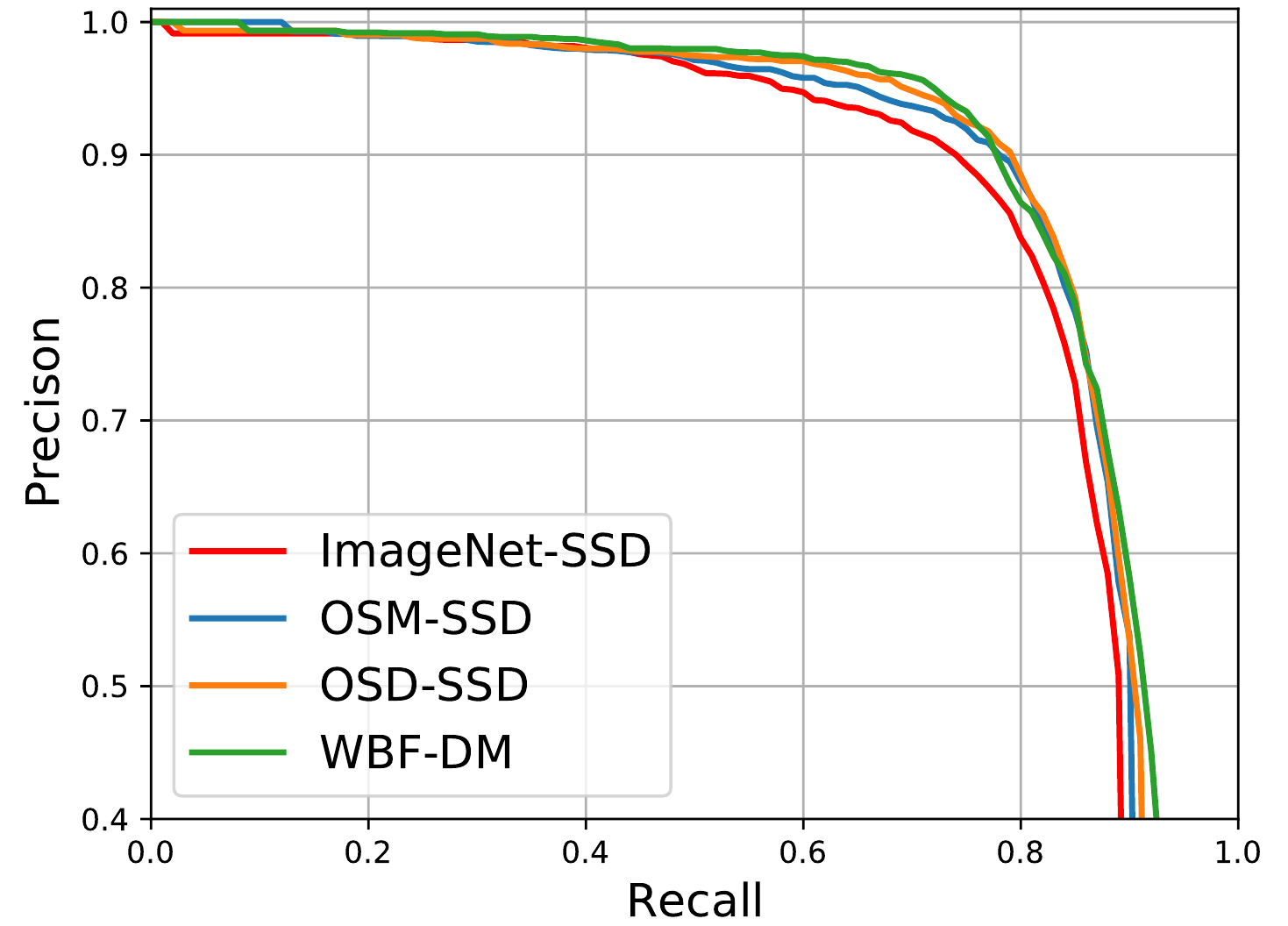}}
	\caption{PR curves of four methods: ImageNet-SSD, OSM-SSD, OSD-SSD and WBF-DM on AIR-SARShip-2.0~\cite{competitation} using the YOLOv3~\cite{redmon2018yolov3} benchmark under $\rm AP_{0.5}$ metric.}
	\label{fig:PR1}
\end{figure}

\begin{figure}[!htb]
	\centering{\includegraphics[scale=0.6]{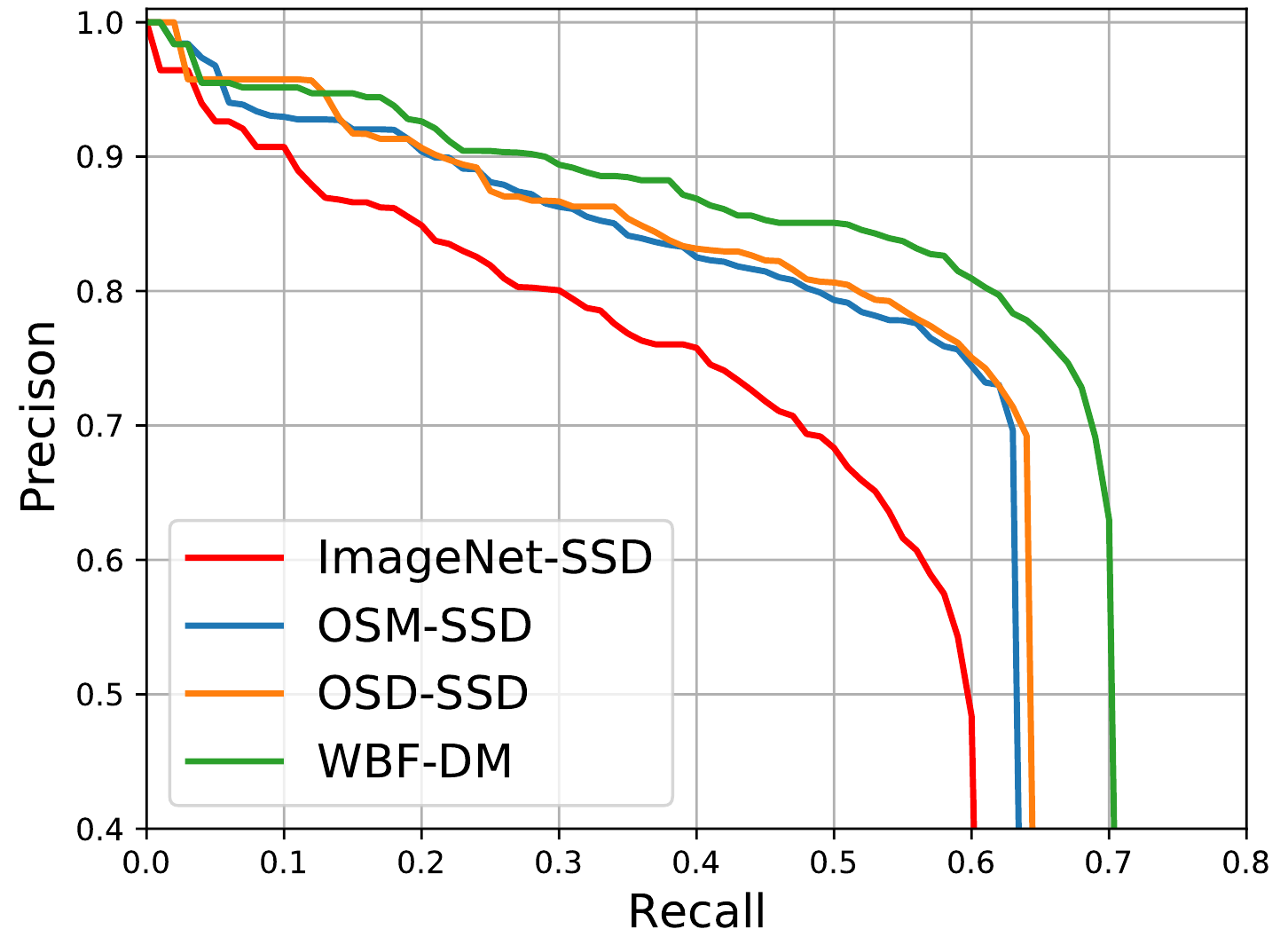}}
	\caption{PR curves of four methods: ImageNet-SSD, OSM-SSD, OSD-SSD and WBF-DM on AIR-SARShip-2.0~\cite{competitation} using the YOLOv3~\cite{redmon2018yolov3} benchmark under $\rm AP_{0.75}$ metric.}
	\label{fig:PR2}
\end{figure}

We use WBF-DM to represent the fusion between predictions of OSD-SSD and OSM-SSD. The final results of WBF-DM on four SAR ship detection results are depicted in Table~\ref{tab:ap1} and Table~\ref{tab:ap2}. Taking the result on AIR-SARShip 2.0~\cite{competitation} using the YOLOv3~\cite{redmon2018yolov3} benchmark as an example, WBF-DM reaches 2.12\% and 1.9\% higher recall rate but 4.8\% and 4.97\% lower precision rates compared to OSD-SSD and OSM-SSD, respectively. This is a natural phenomenon, because WBF-DM inevitably produces false alarms when it preserves as many ships as possible to improve the recall rate. Consequently, $\rm AP_{0.5}$, $\rm AP_{0.75}$ and $\rm AP$ gains a large improvement with a slight decrease in F1 score, which demonstrates that WBF-DM can fully take advantage of complementary characteristics between OSD-SSD and OSM-SSD. 
PR curves of OSD-SSD, OSM-SSD and WBF-DM compared to ImageNet-SSD under $\rm AP_{0.5}$ and $\rm AP_{0.75}$ metrics are depicted in Fig.~\ref{fig:PR1} and Fig.~\ref{fig:PR2}, respectively. It can be observed that the orange curve corresponding to OSD-SSD is always above the red curve corresponding to ImageNet-SSD. Similarly, the blue PR curve denoting OSM-SSD is also above the red PR curve. Although the green curve corresponding to WBF-DM is slightly lower than the orange or red curve at some points in recall rate, the maximum point in recall rate of the green curve is extremely larger than the orange and red curve, leading to the area under the green PR curve increased substantially. This phenomenon also illustrates that WBF is a strategy sacrificing the precision rate for improving the recall rate to achieve a significant increase in $\rm AP$. Finally, WBF-DM yields a considerable increasement in terms of a series of $\rm AP$ metrics compared to ImageNet-SSD, especially 15.34\% improvements in $AP_{0.75}$. All of our experiments manifest positive effects of WBF-DM on improving the comprehensive quality of SAR ship detection.

\label{section:wbf} 
The IoU threshold $thr$ is a significant hyper-parameter in WBF-DM to determine the final fusion results to a large extent. In order to analyze the influence of $thr$, we implement a comparison on AIR-SARShip-2.0~\cite{competitation} using the YOLOv3~\cite{redmon2018yolov3} benchmark with all settings remaining identical except for the value of $thr$ in Table~\ref{tab:wbf1}. It can be observed that when $thr$ turns to be 0.5, $\rm AP_{0.5}$ achieves the highest result of 0.8591. However, as $thr$ gradually increases, $\rm AP_{0.5}$ gradually decreases and $\rm AP_{0.75}$ achieves a maximum value of 0.6188 when $thr$ becomes 0.75. What causes this phenomenon is that $thr$ at the WBF strategy plays NMS's role in the prediction, expecting to match the requirements of localization accuracy. Hence, we can adaptively change the threshold according to different task requirements. In order to achieve a comprehensive performance improvement, we set $thr$ as 0.7 in our experiments.
\renewcommand{\arraystretch}{1.3} 
\begin{table}[!htb]  
	\centering
	\setlength{\tabcolsep}{4mm}  
	\begin{threeparttable} 
		\caption{Results of WBF-DM under different IoU thresholds.}
		\begin{tabular}{c|ccc}
			\hline
			${\rm IoU\ threshold}$&${\rm AP_{0.5}}$&${\rm AP_{0.75}}$&${\rm AP}$ \\
			\hline
			0.4&$0.8533$&$0.6022$&$0.5400$ \cr 
			0.5&$\mathbf{0.8591}$&$0.6086$&$0.5461$ \cr
			0.6&$0.8581$&$0.6141$&$0.5476$ \cr
			0.7&$0.8543$&$0.6156$&$\mathbf{0.5478}$ \cr
			0.75&$0.8502$&$\mathbf{0.6188}$&$0.5466$ \cr
			0.8&$0.8434$&$0.6136$&$0.5424$ \cr
			\hline
		\end{tabular} 
		\label{tab:wbf1} 
	\end{threeparttable}  
\end{table}

\renewcommand{\arraystretch}{1.2} 
\begin{table}[!htb]  
	\centering
	\setlength{\tabcolsep}{1.5mm}   
	\begin{threeparttable} 
		\caption{Fusion results of predictions of ImageNet-SSD, OSM-SSD and OSD-SSD under different combinations.}
		\begin{tabular}{ccc|ccc}
			\hline
			ImageNet-SSD&OSD-SSD&OSM-SSD&${\rm AP_{0.5}}$&${\rm AP_{0.75}}$&${\rm AP}$ \\
			\hline  
			$\surd$&$$&$$&$0.8300$&$0.4622$&$0.4649$ \cr
			 
			$$&$\surd$&$$&$0.8449$&$0.5609$&$0.5096$ \cr
			
			$$&$$&$\surd$&$0.8470$&$0.5509$&$0.5114$ \cr
			
			$\surd$&$\surd$&$$&$0.8496$&$0.5811$&$0.5265$ \cr
			$\surd$&$$&$\surd$&$0.8481$&$0.5842$&$0.5283$ \cr
			$$&$\surd$&$\surd$&$0.8543$&$\mathbf{0.6156}$&$\mathbf{0.5478}$ \cr
			$\surd$&$\surd$&$\surd$&$\mathbf{0.8605}$&$0.6087$&$0.5468$ \cr
			\hline
		\end{tabular} 
		\label{tab:wbf2} 
	\end{threeparttable}  
\end{table}

\renewcommand{\arraystretch}{1.3} 
\begin{table*}[!htb]  
	\centering
	\setlength{\tabcolsep}{3mm}   
	\begin{threeparttable} 
		\caption{Comparison of different CNN-based ship detectors on the SSDD dataset~\cite{li2017ship}.}
		\begin{tabular}{c|c|c|cccccc}
			\hline
			\multicolumn{2}{c|}{Method}
			&Backbone&Precision&Recall&F1&$\rm AP_{0.5}$&$\rm AP_{0.75}$&$\rm AP$ \\
			\hline
			\multirow{5}{*}{One-stage}
			&RetinaNet~\cite{lin2017focal}&ResNet50&$0.8310$&$0.8933$&$0.8610$&$0.8891$&$0.5028$&$-$ \cr
			&RFA-Det~\cite{chen2020r2fa}&ResNet50&$-$&$-$&$-$&$0.9472$&$-$&$-$ \cr
			&DRBox-v2~\cite{an2019drbox}&VGG16&$-$&$-$&$0.9149$&$0.9281$&$-$&$-$\cr
			&FBR-Net~\cite{fu2020anchor}&ResNet50&$0.9279$&$0.9401$&$0.9340$&$0.9410$&$0.5906$&$-$\cr
			\cline{2-9}
			&\textbf{WBF-DM (YOLOv3)}&Darknet53&$\mathbf{0.9413}$&$\mathbf{0.9633}$&$\mathbf{0.9523}$&$\mathbf{0.9577}$&$\mathbf{0.7017}$&$\mathbf{0.6058}$\cr
			\hline  
			\multirow{5}{*}{Two-stage}
			&DCMSNN~\cite{jiao2018densely}&ResNet50&$0.9049$&$0.8914$&$0.8981$&$0.8943$&$0.5417$&$-$ \cr 
			&R-DFPN~\cite{yang2018automatic}&ResNet50&$-$&$-$&$0.8529$&$0.8345$&$-$&$-$ \cr
			&DAPN~\cite{cui2019dense}&ResNet50&$-$&$-$&$-$&$0.8980$&$-$&$-$ \cr
			&HR-SDNet~\cite{wei2020precise}&HRFPN-W40&$-$&$-$&$-$&$0.9730$&$0.7430$&$0.6370$ \cr
			\cline{2-9}
			&\textbf{WBF-DM (Faster R-CNN)}&ResNet50&$\mathbf{0.9079}$&$\mathbf{0.9669}$&$\mathbf{0.9364}$&$\mathbf{0.9740}$&$\mathbf{0.7534}$&$\mathbf{0.6426}$\cr
			\hline
		\end{tabular} 
		\label{tab:sota} 
	\end{threeparttable}  
\end{table*}
In addition to fusing the predictions of OSD-SSD and OSM-SSD, we also conduct experiments to fuse detection results of different combinations among ImageNet-SSD, OSD-SSD and OSM-SSD. The fusion results are shown in Table~\ref{tab:wbf2}. We can see that, regardless of any combination, the performance of fusing multiple models is always superior to that of a single model, demonstrating the effectiveness of the WBF strategy. Furthermore, the fusion of OSD-SSD and OSM-SSD outperforms not only the fusion of ImageNet-SSD and OSD-SSD but also the fusion of ImageNet-SSD and OSM-SSD. Compared to ImageNet-SSD fused with OSD-SSD, the higher improvements can be contributed to the fewer false alarms on land area of OSM-SSD. Compared to ImageNet-SSD combined with OSM-SSD, the improved detection results are due to the higher recall on sea area of OSD-SSD. It is the complementary advantages under different scenes that allow our WBF strategy between OSD-SSD and OSM-SSD to significantly improve the ship detection performance. Finally, when we combine the predictions of ImageNet-SSD, OSD-SSD and OSM-SSD, $\rm AP_{0.5}$ achieves the highest value of 86.05\% while $\rm AP_{0.75}$ and $\rm AP$ become lower. In other words, the improvements vanish as the requirement of localization accuracy gets higher. We conjecture that instead of complementary advantages in different scenes, the increased $\rm AP_{0.5}$ is because the WBF strategy adjusts not only the confidence of the predicted boxes, but also the position, reaching a better localization accuracy.

\subsubsection{Comparison With CNN-Based SAR ship detectors}
We split CNN-based SAR ship detectors into two categories: one-stage ship detector and two-stage ship detector for a fair comparison. The overall detection results of different detectors on the SSDD dataset~\cite{li2017ship} are listed in Table~\ref{tab:sota}. For one-stage detector, we compare the proposed WBF-DM method based on the YOLOv3~\cite{redmon2018yolov3} benchmark with four CNN-based methods including RetinaNet~\cite{lin2017focal}, RFA-Det~\cite{chen2020r2fa}, DRBox-v2~\cite{an2019drbox} and FBR-Net~\cite{fu2020anchor}. Specifically, our method achieves 1.34\% and 2.32\% higher precision rate and recall rate compared to the FBR-Net method, respectively. In terms of F1 score and $\rm AP_{0.5}$, our method obtains 1.83\% and 1.67\% higher improvements. Furthermore, our method considerably improves $\rm AP_{0.75}$ by a large margin of 11.11\%, indicting that the predicted boxes with a higher localization accuracy can be obtained. Compared to other methods, similar enhancements can be also achieved. As for two-stage detector, we also compare our WBF-DM based on the Faster R-CNN~\cite{ren2015faster} benchmark with four CNN-based methods including DCMSNN~\cite{jiao2018densely}, R-DFPN~\cite{yang2018automatic}, DAPN~\cite{cui2019dense} and HR-SDNet~\cite{wei2020precise}. It is observed that our WBF-DM method can achieve considerably improvements in all degrees compared to other state-of-the-art ship detectors. More importantly, our methods can be easily combined with other detectors to further boost SAR ship detection. All of these improvements verify the effectiveness and complementarity of OSD-SSD and OSM-SSD, and the superiority of the combination of the two detectors.

\section{Conclusion}
Considering directly leveraging ImageNet pretraining technique as common used is hardly to obtain a good SAR ship detector, this paper introduced a completed framework to boost ship detection performance in SAR images. Specifically, to resolve problems that ships from ImageNet are different from ships from earth observations, we first proposed an OSD pretraining technique to improve the feature learning ability of ships in SAR images by fully taking advantage of ships’ annotation information from aerial images. Secondly, to handle the problem of different imaging geometry between optical and SAR images, we proposed an OSM pretraining technique to enhance the general feature embedding of SAR images by common representation learning via bridge neural networks. Thirdly, observing the different advantages of OSM-SSD and OSD-SSD, this paper employed the WBF strategy to combine the predictions of OSD-SSD and OSM-SSD to further imporve detection results in SAR images. Finally, various experiments have been conducted on four SAR ship detection datasets and two representative CNN-based detection benchmarks to verify the effectiveness and complementarity of OSD-SSD and OSM-SSD, and the state-of-the-art performance of the combination of the two detectors. In the future, we will consider explore the performance of the proposed method on more complicated networks and more challenging datasets.

\bibliographystyle{IEEEbib}
\bibliography{egbib}

\begin{IEEEbiography}
	[{\includegraphics[width=1in,height=1.25in,clip,keepaspectratio]{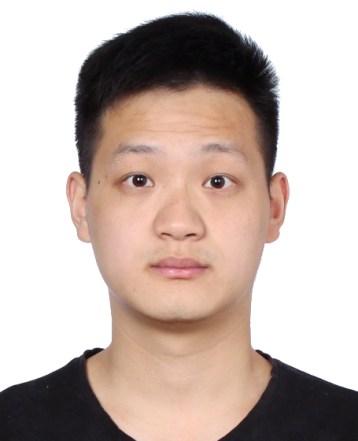}}]{Wei Bao} received a B.S. degree from Nanjing Tech University, Nanjing, China, in 2018. He is currently pursuing the master’s degree with Beijing Institute of Technology, Beijing, China and performing cooperation research with researchers at the Qian Xuesen Laboratory of Space Technology, China Academy of Space Technology, Beijing, China. 
	His research interests include computer vision, few-shot learning and remote sensing object detection.
\end{IEEEbiography}

\begin{IEEEbiography}
	[{\includegraphics[width=1in,height=1.25in,clip,keepaspectratio]{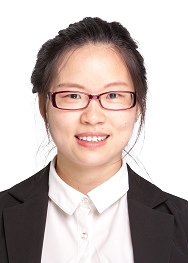}}]{Meiyu Huang} received a B.S. degree in computer science and technology from Huazhong University of Science and Technology, Wuhan, China, in 2010, and a Ph.D. degree in computer application technology from the University of Chinese Academy of Sciences, Beijing, China, in 2016. She is currently an assistant researcher in the Qian Xuesen Laboratory of Space Technology, China Academy of Space Technology, Beijing, China. 
	Her research interests include machine learning, ubiquitous computing, human-computer interaction, computer vision and image processing.
\end{IEEEbiography}

\begin{IEEEbiography}
	[{\includegraphics[width=1in,height=1.25in,clip,keepaspectratio]{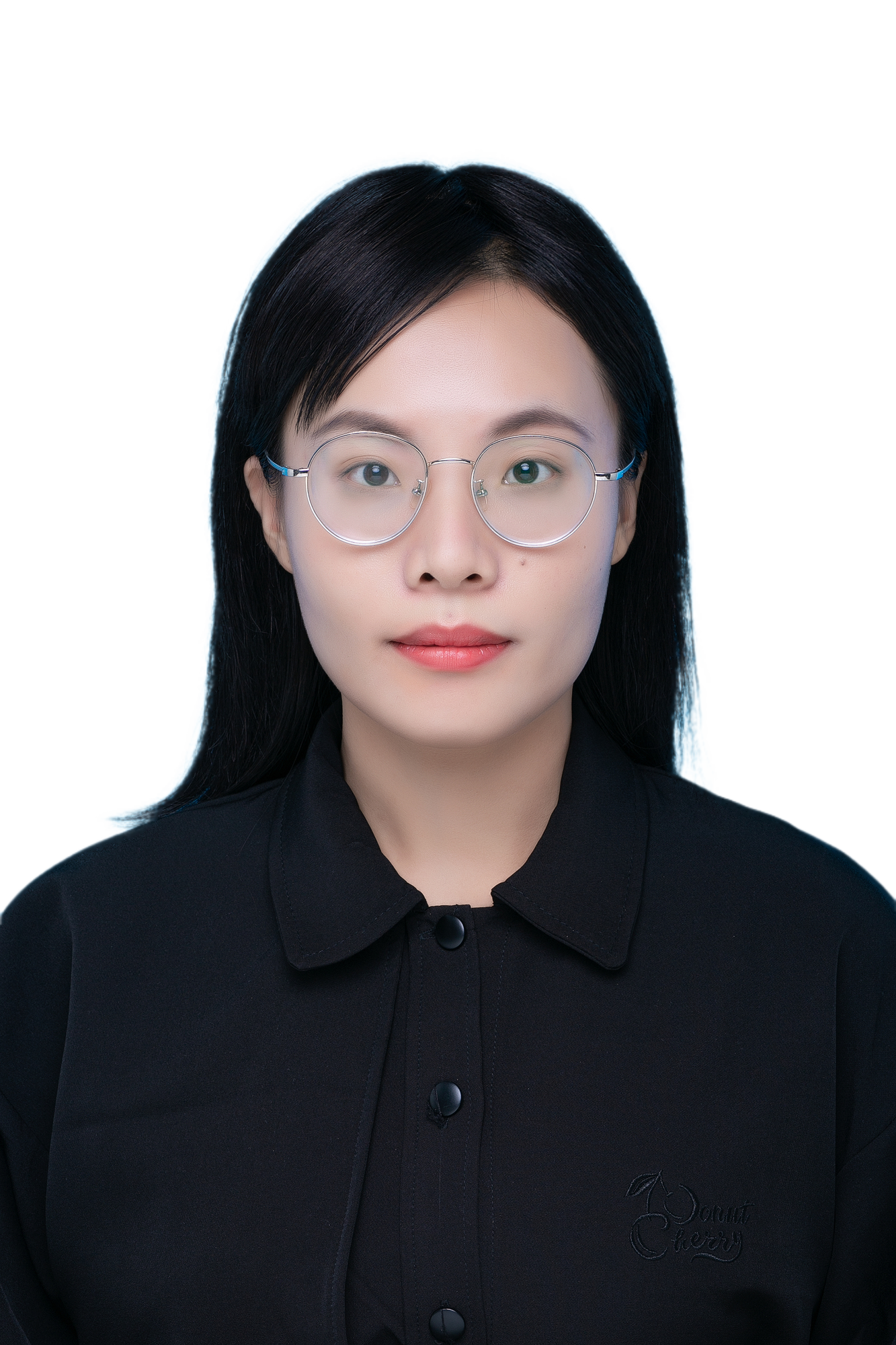}}]{Yaqin Zhang} received the B.S. degree from the School of Science, North University of China, Taiyuan, China, in 2014. She is currently pursuing the master’s degree with School of Mathematics and Computational Science, Xiangtan University, Xiangtan. Since 2020, she has performing cooperation research with researchers at the Qian Xuesen Laboratory of Space Technology, China Academy of Space Technology, Beijing, China. 
	Her current research interests include the area of computer vision and computational imaging.
\end{IEEEbiography}

\begin{IEEEbiography}
	[{\includegraphics[width=1in,height=1.25in,clip,keepaspectratio]{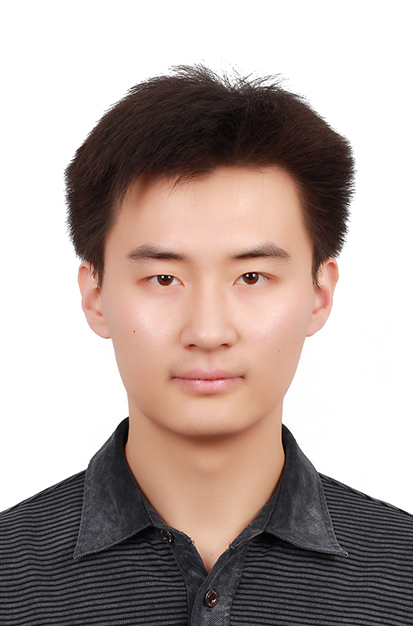}}]{Yao Xu} received a B.S. degree in electrical and computer engineering in Shanghai Jiao Tong University, China, in 2013 and a M.S. degree in electrical and computer engineering in University of California, Irvine, US, in 2016. He is currently an assistant researcher in the Qian Xuesen Laboratory of Space Technology, China Academy of Space Technology, Beijing, China. 
	His research interests include deep learning, data Fusion, distributed system and computer architecture.
\end{IEEEbiography}

\begin{IEEEbiography}
	[{\includegraphics[width=1in,height=1.25in,clip,keepaspectratio]{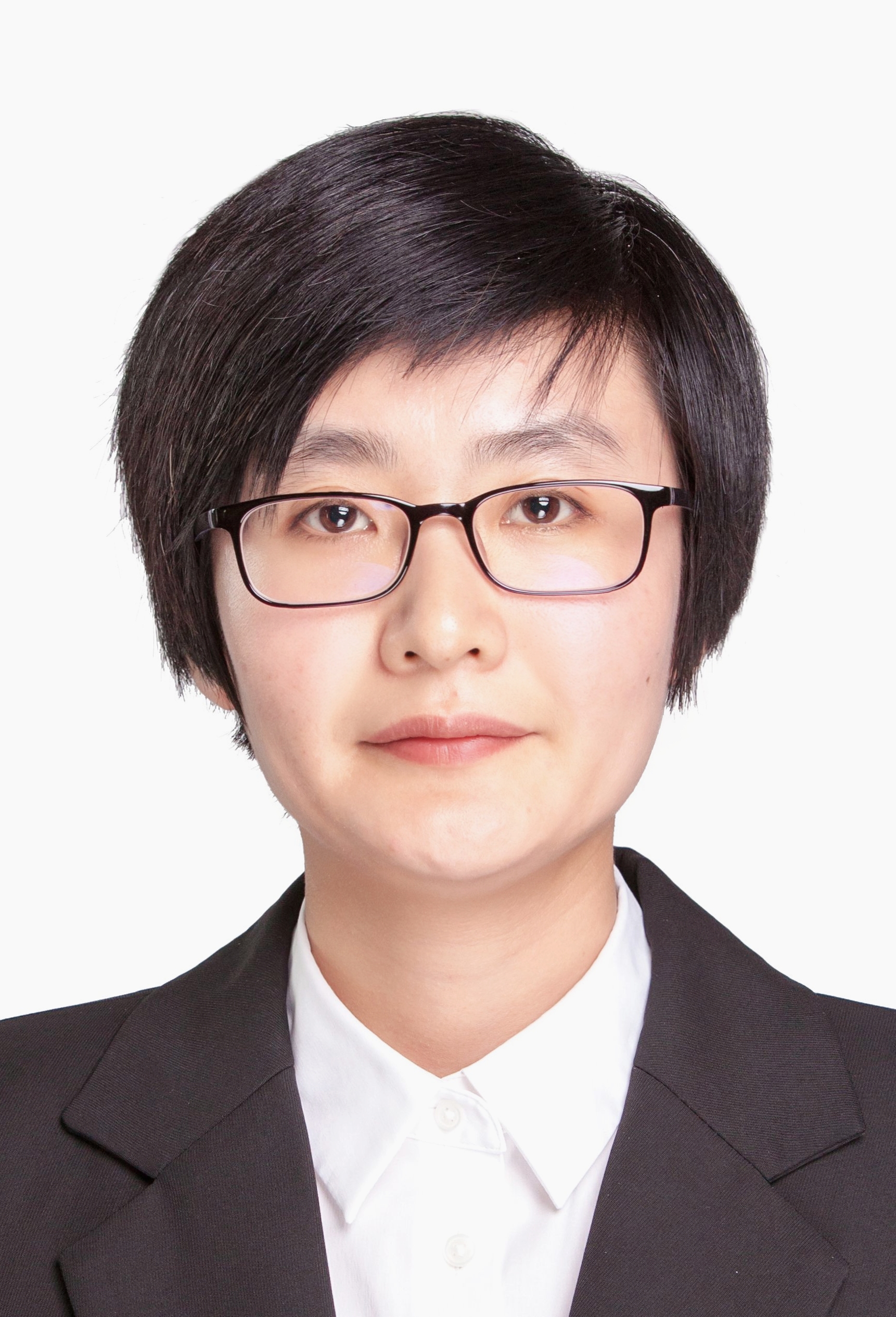}}]{Xuejiao Liu} received the B.S. degree from Jilin University, Changchun, China, in 2013, and the Ph.D. degree in computational mathematics from University of Chinese Academy of Sciences, Beijing, China, in 2018. She is currently an assistant researcher in the Qian Xuesen Laboratory of Space Technology, China Academy of Space Technology, Beijing, China. Her research interests include deep learning, generative model, and numerical simulation.
\end{IEEEbiography}

\begin{IEEEbiography}
	[{\includegraphics[width=1in,height=1.25in,clip,keepaspectratio]{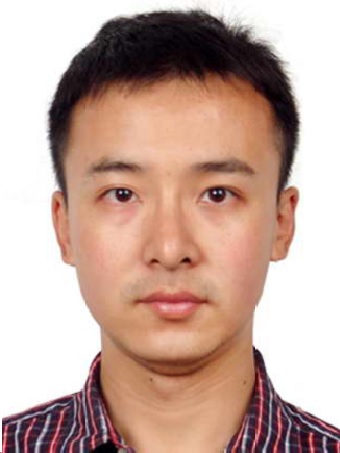}}]{Xueshuang Xiang} received a B.S. degree in computational mathematics from Wuhan University, Wuhan, China, in 2009, and a Ph.D. degree in computational mathematics from the Academy of Mathematics and Systems Science of Chinese Academy of Sciences, Beijing, China, in 2014. In 2016, he was a postdoctoral researcher with the Department of Mathematics, National University of Singapore, Singapore. He is currently an associate researcher in the Qian Xuesen Laboratory of Space Technology, China Academy of Space Technology, Beijing, China. 
	His research interests include numerical methods for partial differential equations, image processing and deep learning.
\end{IEEEbiography}
\vfill

\end{document}